\documentclass{article}

\usepackage{PRIMEarxiv}

\usepackage{minted}
\usepackage[utf8]{inputenc} 
\usepackage[T1]{fontenc}    
\usepackage{hyperref}       
\usepackage{url}            
\usepackage{booktabs}       
\usepackage{amsfonts}       
\usepackage{nicefrac}       
\usepackage{microtype}      
\usepackage{lipsum}
\usepackage{fancyhdr}       
\usepackage{graphicx}       
\graphicspath{{images/}}     
\usepackage{amsmath}

\title{Leveraging an Atmospheric Foundational Model for Subregional Sea Surface Temperature Forecasting}

\author{
  \begin{minipage}[t]{0.32\textwidth}
        \centering
        Víctor Medina \\
        {\normalfont Centro de Tecnologías de la Imagen (CTIM)\\
        Instituto Universitario de Cibernética, Empresas y Sociedad (IUCES)\\
        University of Las Palmas de Gran Canaria, Spain\\
        \texttt{victor.medina110@alu.ulpgc.es}}
  \end{minipage}
  \hfill  
  \begin{minipage}[t]{0.32\textwidth}
        \centering
        Giovanny A. Cuervo-Londoño\\
        {\normalfont Oceanografía Física y Geofísica Aplicada (OFYGA) \\
        Instituto Universitario de Investigación en Acuicultura Sostenible y Ecosistemas Marinos (ECOAQUA)\\
        University of Las Palmas de Gran Canaria, Spain\\
        \texttt{giovanny.cuervo101@alu.ulpgc.es}}
  \end{minipage}
  \hfill
  \begin{minipage}[t]{0.32\textwidth}
        \centering
        Javier Sánchez\\
        {\normalfont Centro de Tecnologías de la Imagen (CTIM) \\
        Instituto Universitario de Cibernética, Empresas y Sociedad (IUCES)\\
        University of Las Palmas de Gran Canaria, Spain\\
        \texttt{jsanchez@ulpgc.es}}
  \end{minipage}
}

\begin{document}
\maketitle

\begin{abstract}
The accurate prediction of oceanographic variables is crucial for understanding climate change, managing marine resources, and optimizing maritime activities. Traditional ocean forecasting relies on numerical models; however, these approaches face limitations in terms of computational cost and scalability. In this study, we adapt Aurora, a foundational deep learning model originally designed for atmospheric forecasting, to predict sea surface temperature (SST) in the Canary Upwelling System. By fine-tuning this model with high-resolution oceanographic reanalysis data, we demonstrate its ability to capture complex spatiotemporal patterns while reducing computational demands. Our methodology involves a staged fine-tuning process, incorporating latitude-weighted error metrics and optimizing hyperparameters for efficient learning. The experimental results show that the model achieves a low RMSE of 0.119K, maintaining high anomaly correlation coefficients (ACC $\approx 0.997$). The model successfully reproduces large-scale SST structures but faces challenges in capturing finer details in coastal regions. This work contributes to the field of data-driven ocean forecasting by demonstrating the feasibility of using deep learning models pre-trained in different domains for oceanic applications. Future improvements include integrating additional oceanographic variables, increasing spatial resolution, and exploring physics-informed neural networks to enhance interpretability and understanding. These advancements can improve climate modeling and ocean prediction accuracy, supporting decision-making in environmental and economic sectors.
\end{abstract}


\section{Introduction}

The ocean plays a crucial role in multiple aspects of the planet, from regulating the global climate~\cite{ipcc_ar6_2021} to the availability of strategic marine resources \cite{fao2022state_of_fisheries}. Understanding its processes and predicting changes is essential for decision-making in the fishing industry and maritime transport sectors~\cite{chassignet2006modeling}. Anticipating the behavior of oceanographic variables is important in the current scenario~\cite{etoffs2022challenges}, marked by climate change and the urgent need for sustainable policies~\cite{un_ocean_agenda2030_2022}, including the development of new tools to assess the present and future states of the Earth, as current approaches are still limited~\cite{zhao2024applications}.

Ocean analysis is critical since it influences global climate, storm formation, and heat exchange with the atmosphere through sea surface temperature~\cite{foxkemper2019challenges}. Additionally, marine biodiversity depends on the dynamic balance of the ocean's physical and chemical properties. Human activities also require information about ocean conditions to ensure safety, optimize maritime routes, manage energy resources, and implement ecological policies~\cite{noaa_ocean_policy2021}.

Traditional ocean prediction methodologies rely on numerical models that solve complex fluid dynamics equations. However, these models struggle to capture data across multiple spatial and temporal scales and have high computational costs. These limitations, combined with the increasing volume of data, create new research opportunities. In this context, deep learning has emerged as a promising alternative to overcome these challenges, enabling the processing of large volumes of data across different scales. Furthermore, with the growing quantity and diversity of data from \textit{in situ} sources, satellites, and reanalysis, we now have an opportunity to address these challenges~\cite{EMB2020bigdata}.

These new models learn complex relationships from data without explicitly relying on physical equations. By removing the strict dependence on theoretical formulations, it becomes feasible to capture patterns that emerge from the interaction of multiple variables. This approach can provide more detailed solutions, reduce computational costs, and integrate heterogeneous data. Models based on deep learning could enable the development of more efficient methods in various scenarios~\cite{wmo_ocean2021}.

The field of atmospheric prediction has witnessed a transformative shift with the advent of deep learning models. Recent breakthroughs include models like FourCastNet~\cite{pathak2022fourcastnet}, known for its speed and accuracy in predicting extreme weather events, and GraphCast~\cite{lam2022graphcast}, which demonstrates remarkable skill in medium-range global weather forecasting based on Graph Neural Networks. Furthermore, models such as NeuralGCM~\cite{kochkov2024neural} explore the integration of neural networks within general circulation models, aiming to enhance the representation of sub-grid scale processes. More recently, the Aurora~\cite{bodnar2024aurora} model has emerged, showcasing the potential of deep learning to not only predict atmospheric conditions but also to extend these capabilities to coupled systems by learning complex interactions across different Earth system components. These advancements signify a new era in weather and climate modeling, promising more efficient and accurate predictions for a range of applications.

This work aims to adapt the Aurora foundational model, initially developed for global atmospheric applications, to subregional ocean prediction. The model is fine-tuned for predicting the evolution of the SST in the Canary Upwelling System. We are interested in adapting the model for predicting the ocean’s potential temperature ($\theta_0$) in a specific geographic area and period. Its three-dimensional architecture enables the efficient integration of complex and heterogeneous data. Additionally, its reduced training time, ability to leverage previously learned representations, and flexibility to incorporate data from various sources make it an ideal candidate for tackling current challenges in ocean prediction. 

We use data from the GLORYS12V1~\cite{lellouche2018copernicus} ocean prediction system of the Copernicus Marine Service~\cite{copernicus}. This corresponds to the global ocean physical reanalysis, which provides three-dimensional daily fields with a horizontal resolution of 1/12° and 50 vertical levels. This dataset, based on the assimilation of satellite and \textit{in situ} observations, provides information on potential temperature, salinity, currents, and two-dimensional variables, such as sea level and mixed-layer thickness.

This model offers advantages in terms of computational cost reduction and prediction quality, although limitations exist due to the complexity of the domain and physical constraints. This should be seen as complementary to other methodologies rather than a complete replacement. 

Section~\ref{se:relate_work} summarizes state-of-the-art works. Section~\ref{se:dataset} details the dataset used in this work and the area of study. Section~\ref{se:method} explains the Aurora model and how we adapted it to our problem. The experimental configuration in Section~\ref{se:experimental_settup} addresses various details about the training process, the loss function, and the metrics used to evaluate the model. Section~\ref{se:results} tackles the fine-tuning process and shows some preliminary results. Finally, the conclusions in Section~\ref{se:conclusion} summarize the main ideas and contributions of the work and propose some ideas for future research.

\section{Related Work}
\label{se:relate_work}

Prediction in oceanography is essential for anticipating and understanding natural phenomena that affect global climate and various human activities~\cite{lehodey2006climate}. Sea temperature plays a fundamental role in the distribution of water masses, global circulation, and the ocean-atmosphere energy balance~\cite{robinson2010coastal, mcphaden2006enso}. Thus, estimating changes in the physical and chemical properties of the ocean, such as temperature, is crucial for multiple activities~\cite{rs16030504}, in addition to being key to anticipating natural risks.  

Ocean observation and prediction results from global efforts~\cite{wmo_ocean2021}, in which multiple national and international agencies contribute scientific data. Among them, the World Meteorological Organization (WMO) stands out as it establishes frameworks for international cooperation in data exchange, standardization of procedures, and guidelines that facilitate climate understanding on a planetary scale~\cite{wmo2020}.

The Copernicus Program, promoted by the EU in collaboration with the ESA, provides services and information about the  ocean~\cite{copernicus}. The Copernicus Marine Environment Monitoring Service (CMEMS) integrates satellite observations, \textit{in situ} measurements, and numerical models to offer various oceanographic products. This service supplies data on sea temperature, salinity, currents, and sea level variables. As a result, it provides a solid foundation for scientific studies, operational applications, and decision-making in marine environments~\cite{vonSchuckmann2018}. On the other hand, the USA National Oceanic and Atmospheric Administration (NOAA) also collects and processes oceanic and atmospheric data. These data and products have become standard references for the international community~\cite{noaa}. Additionally, ECMWF provides reanalysis and numerical models, globally recognized for their quality and accuracy~\cite{ecmwf}.

Satellite observations have revolutionized ocean monitoring. Missions such as ESA and Copernicus' Sentinel-3 or NASA's MODIS~\cite{modis} allow for detailed information on sea surface temperature, sea level height, surface salinity, and chlorophyll concentration---among other variables---characterized by high spatial and temporal resolution. However, satellite measurements can be affected by clouds or other atmospheric factors. To compensate for these shortcomings and obtain a more comprehensive view, it is necessary to combine these observations with other data sources~\cite{donlon2009successes}.

\textit{In situ} data come from direct measurements taken by buoys, research vessels, fixed platforms, and autonomous underwater vehicles~\cite{argo2000}. These measurements provide precise data on key variables. Although these measurements have limited spatial and temporal coverage, their high quality and detail are indispensable for calibration, validation, and assimilation in numerical models.  Reanalysis and numerical models integrate satellite and \textit{in situ} data, offering coherent representations of the ocean. 

Numerical Weather Prediction (NWP) models have been essential for predicting ocean dynamics for decades. These models are based on fluid dynamics and thermodynamics equations, solving the evolution of fundamental variables over three-dimensional grids that cover the globe~\cite{kalnay2003atmospheric}. The main advantage of NWP lies in its physical consistency, as it is based on fundamental laws governing ocean motion. However, it also has some limitations, such as the high computational cost required to make predictions, which makes high-resolution simulations expensive. Additionally, the complexity of ocean-atmosphere interactions leads to small errors that accumulate over time, reducing the accuracy of long-term predictions and limiting the ability to capture small-scale phenomena~\cite{bauer2015quiet}.  

These challenges have led to methodologies aimed at increasing the capacity to process large volumes of data in a computationally scalable manner~\cite{liu2024deep}. In this regard, deep learning has emerged as a promising approach to improving efficiency and accuracy when working with heterogeneous data.

Initially, CNNs~\cite{lecun1995convolutional} gained popularity in meteorology and oceanography due to their ability to capture spatial patterns in two-dimensional data such as temperature maps or ocean currents. Some use cases included precipitation prediction and identifying spatial patterns in the ocean~\cite{shi2015convolutional}. However, these models were inefficient at capturing temporal dependencies or three-dimensional structures, limiting their effectiveness for more complex tasks.  

Later, with the emergence of graph neural networks (GNN), GraphCast \cite{lam2022graphcast} was introduced in 2023, a model capable of tackling medium-range global weather prediction with high detail. This approach leverages graph structures to capture complex spatial relationships, showing promising results in forecasting meteorological and climate variables. Additionally, diffusion-based methods such as GenCast~\cite{du2023gencast} were developed to integrate uncertainty into the models. This stochastic approach generates multiple future states, making it a valuable tool for scenario analysis and risk assessment, particularly in capturing ocean variability. The Artificial Intelligence/Integrated Forecasting System (AIFS)~\cite{lang2024aifsecmwfsdatadriven} is an evolution of GraphCast in which the processor is replaced with a shifted window attention Transformer. 

GNNs have been adapted to ocean prediction. For instance, SeaCast~\cite{holmberg2024regional} implements medium-range ocean forecasting in a limited area region defined by the Mediterranean Sea, and a more recent approach~\cite{londono2025forecasting,clondonio2025deep} predicts the SST in the region of the Canary Islands and the Northwest African coast. The importance of the underlying mesh in these architectures has been studied in ~\cite{reyes2025adaptive}: Structured meshes produce artifacts in medium-range forecasting, which can be tackled using bathymetry-aware unstructured meshes, as shown in~\cite{londono25voronoi}.

One of the most significant breakthroughs in this trajectory came with Transformers~\cite{vaswani2017attention}. Their attention mechanism enabled a more efficient representation of spatial and temporal relationships. FourCastNet~\cite{pathak2022fourcastnet} integrated adaptive Fourier operators to perform fast and accurate global climate predictions while reducing computational complexity. The three-dimensional architecture of Pangu-Weather~\cite{bi2022panguweather3dhighresolutionmodel} improved vertical representation and the ability to capture complex atmospheric phenomena.  

Over time, Transformers were further adapted for oceanography. For instance, XiHe~\cite{zhuang2021xihi}, a model designed for global ocean eddy-resolving forecasting, combined the GLORYS12 reanalysis with data from ERA5~\cite{bell2021era5} and satellite observations to anticipate the evolution of physical variables, providing high-resolution estimates. Another example is Orca~\cite{guo2024orca}, which employed an encoder-based architecture, a fusion module, and a decoder with temporal attention mechanisms to simulate global ocean circulation.  

More recently, a three-dimensional foundational model, Aurora~\cite{bodnar2024aurora}, combined pretraining on vast amounts of heterogeneous data with a fine-tuning process tailored to different scales and variables. Inspired by successes in other fields, such as protein structure prediction and natural language processing~\cite{abramson2024protein, bommasani2021openaifoundation}, Aurora is capable of generating high-resolution forecasts with lower computational costs and greater accuracy in extreme events. This model stands out for its ability to handle sparse and heterogeneous data while overcoming the limitations of other architectures by integrating with different types of climate data.

Table \ref{tab:comparacion_modelos_dl} illustrates the evolution of deep learning in climate and oceanographic prediction over time. Organized chronologically, this table helps us understand improvements in resolution, the type and quantity of variables that can be handled, and whether the approach is deterministic (D) or stochastic (E).

\begin{table}[ht!]
\caption{Table with a chronological comparison and key features of various deep learning models applied to climate and oceanic prediction. In the last column, (D) stands for Deterministic and (S) for Stochastic}
\begin{tabular}{l l l l l}
\hline
Model & Year & Resolution & Number of Variables & Forecasting \\
\hline
CNN & 2010s & Medium & Limited & D \\
RNN/LSTM & 2010s & Medium & Limited & D \\
XiHe (Transformer) & 2021 & High & Large & D \\
FourCastNet (Transformer) & 2022 & High & Large & D \\
Pangu-Weather (Transformer 3D) & 2022 & High & Large & D/S \\
GraphCast (GNN) & 2023 & High & Large & D \\
GenCast (Diffusion model) & 2023 &  High & Large & S \\
AIFS (GNN) & 2023 &  High & Large & D/S \\
Orca (Transformer) & 2024 & High & Various & D \\
Aurora (Transformer) & 2024 & High & Large & D \\
\hline
\end{tabular}
\label{tab:comparacion_modelos_dl}
\end{table}

\section{Dataset and Study Area}
\label{se:dataset}
This section presents the data used in this work, including the study area, period, and selected variables. It details the data source, the rationale for choosing the Northeastern Atlantic region, and the time frame. The complexity of patterns in the study area is also examined. Finally, it explains the temporal division of data for training, validation, and testing.

\subsection{Global Ocean Physics Reanalysis Data}

The CMEMS~\cite{CMEMS2018} is a European service dedicated to monitoring the marine environment, providing a wide range of global and regional oceanic data for the scientific community. Its products are organized into different thematic categories: the Blue Ocean, which focuses on physical variables such as temperature, salinity, sea level, and currents; the Green Ocean, which covers biogeochemical variables like chlorophyll concentration and dissolved oxygen; and the White Ocean, which includes data on sea ice and polar regions. This classification enables users to filter information based on specific project needs, ensuring high-quality and relevant data~\cite{copernicus}.

In this work, we selected the Blue Ocean category, specifically the GLOBAL\_MULTIYEAR\_PHY\_001\_030 product, which is formulated as a global physical ocean reanalysis. This product spans from 1993 to 2021, offering data with a spatial resolution of $1/12^o$ and fifty vertical levels. The GLORYS12V1 ocean prediction system is part of the reanalysis operated by Mercator Ocean, based on the NEMO model. This scheme integrates techniques such as SEEK to assimilate satellite data (sea level altimetry, sea surface temperature) and \textit{in situ} observations (ARGO profiles and oceanographic cruises), achieving a coherent and physically consistent representation of the global ocean environment~\cite{Madec2015, Lellouche}. The primary function of this service is to offer reliable and up-to-date information on ocean conditions. 

The product includes multiple data configurations: daily and monthly averages of three-dimensional variables such as potential temperature, salinity, and currents, as well as surface variables like sea level height, mixed layer thickness, and sea ice characteristics. This structure allows the analysis of mesoscale variability and seasonal patterns, providing a versatile dataset suitable for dynamic and climate studies.  

The data is available in NetCDF format, which follows the CF-1.6 convention, ensuring standardization of variables and metadata~\cite{Rew1990, CMEMSManual2024}.  

The selection of this product is driven by the need for a dataset that combines high spatial resolution with extensive temporal coverage and rigorous scientific quality~\cite{CMEMS2018}. This approach enables the capture of key dynamic phenomena, such as ocean eddies and fronts, while also allowing for the analysis of long-term trends and seasonal cycles.

\subsection{Area and Period of Study}

Our study area is limited to the Northeastern Atlantic region, near the African coast and off the Canary Islands, an area of interest due to its intense coastal upwelling phenomena. This region is characterized by high biological productivity and a significant influence on the regional climate~\cite{vazquez2021assessment}. 

\begin{figure}[ht!] 
\centering 
\includegraphics[width=0.6\textwidth]{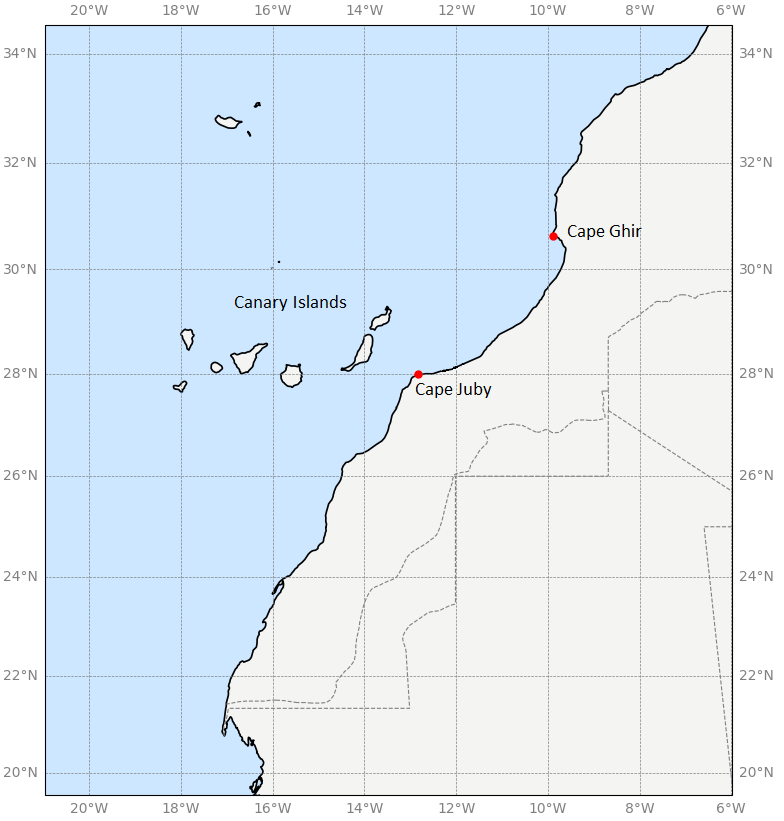} 
\caption{Study area showing the African coast, the Canary Islands, and relevant capes.} 
\label{fig:mapa_zona_estudio} 
\end{figure}

Figure~\ref{fig:mapa_zona_estudio} shows a map of the selected area, highlighting key geographic features relevant to coastal upwellings, such as Capes Ghir and Juby. These capes are crucial in coastal currents and nutrient transport to the surface, promoting phytoplankton proliferation~\cite{mcphaden2006enso}. Tabla~\ref{tab:parametros} shows the coordinates that define our area of interest.

\begin{table}[ht!]
\caption{Geographic coordinates of the study area.}
\centering
\begin{tabular}{|l|c|}
\hline
\textbf{Parameter}       & \textbf{Value}       \\ \hline
Minimum Longitud         & $-20.97^\circ$       \\ \hline
Maximum Longitud         & $-5.975^\circ$       \\ \hline
Minimum Latitud          & $19.55^\circ$        \\ \hline
Maximum Latitud          & $34.525^\circ$       \\ \hline
\end{tabular}
\label{tab:parametros}
\end{table}

This region is characterized by a strong influence of the trade winds and the Canary Current, both considered main factors of coastal upwelling, affecting biodiversity and local dynamics~\cite{rs16030504}. Our data period extends from January 1, 2014, to January 1, 2021. This seven-year interval allows the exploration of seasonal variations and medium-term trends.

\subsection{The Potential Temperature}

Our dataset contains a series of variables that provide a general overview of ocean dynamics. These variables range from potential temperature to salinity, sea level height, mixed layer depth, and zonal and meridional components of current velocity~\cite{foxkemper2019challenges}.

\begin{figure}[ht!] 
\centering 
\includegraphics[width=\textwidth, height=7cm]{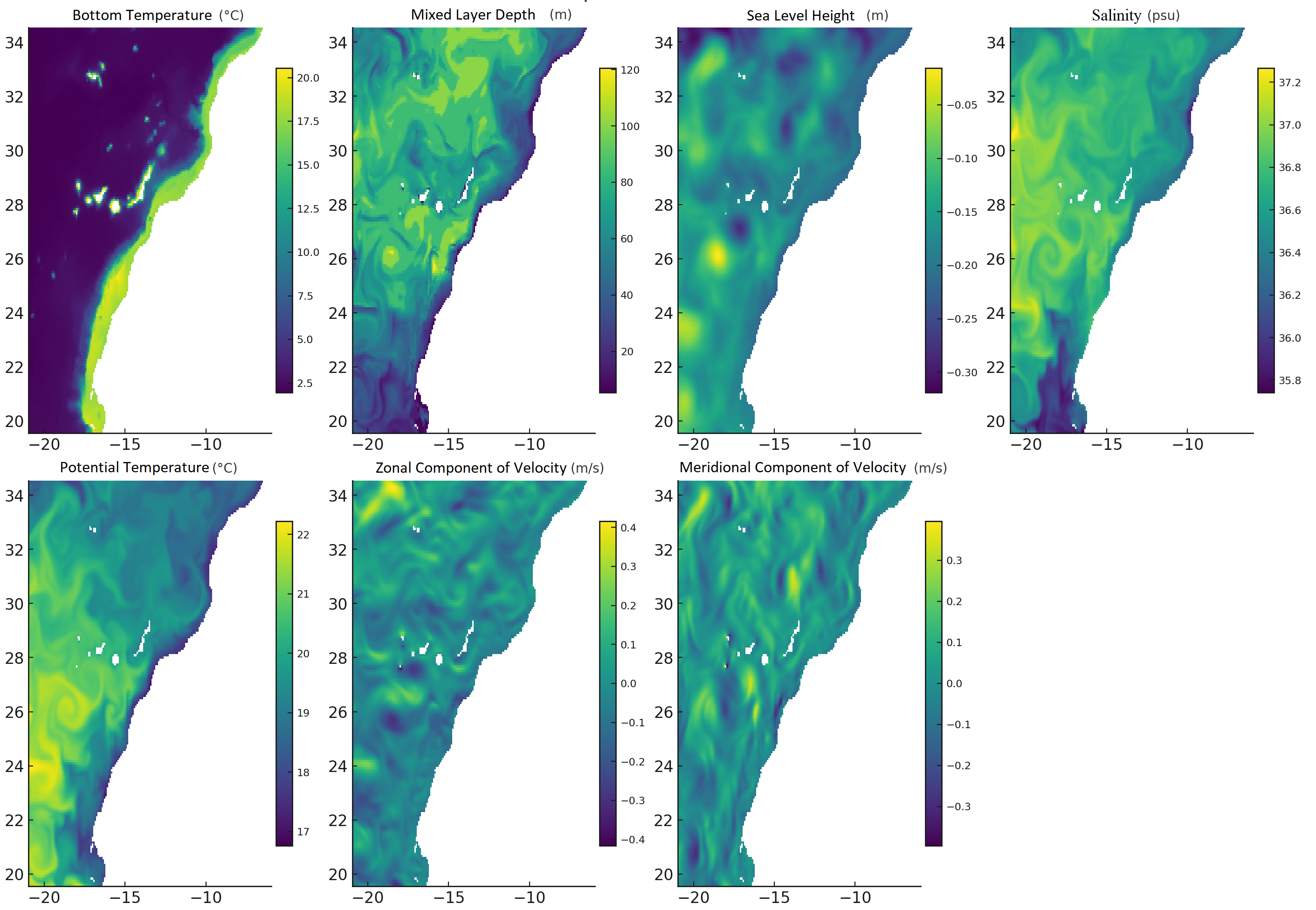} 
\caption{Spatial representations of the dataset variables from the GLORYS12V1 product.} 
\label{fig:ejemplo_subplots} 
\end{figure}

Figure~\ref{fig:ejemplo_subplots} shows a set of maps for each variable, representing the region of interest. Different patterns highlight the influence of the trade winds and coastal upwelling, showing these factors as generators of distinct thermal and saline gradients.

In this work, we use the potential temperature, $\theta_0$. It is defined as the temperature of the water brought to atmospheric pressure. This concept eliminates the effects that pressure generates at different depths. Hence, potential temperature becomes a fundamental tool for comparing water masses and studying the thermal structure of the ocean~\cite{talley2011descriptive}. The $\theta_0$ map shows that the warmest waters are located in the south at the open ocean, while the coldest waters are found in the north and the African coast, in the upwelling region. This pattern is influenced by coastal upwelling and the interactions between the Canary Current and the trade winds~\cite{rs16030504}. The $\theta_0$ map also shows thermal gradients along the coast, indicating active mixing processes. This behavior affects marine biodiversity, as colder waters transport nutrients from the depths to the surface~\cite{zhao2024applications}.

In addition to its relevance in thermal structure, $\theta_0$ is essential for understanding heat exchange processes between the ocean and the atmosphere~\cite{robinson2010coastal}, thus, it becomes a fundamental tool for analyzing seasonal evolution and its impact on the ocean~\cite{mcphaden2006enso}.

On the other hand, salinity shows a higher concentration in the open ocean, possibly due to thermal influence and ocean circulation~\cite{foxkemper2019challenges}. Sea level height shows patterns of mesoscale dynamics, evidencing depressions and elevations characteristic of ocean eddies~\cite{liu2024deep}. Mixed layer depth shows low values near the coast, reflecting the influence of surface waters and upwelling masses.

The integration of these variables allows for a holistic view of ocean dynamics in this region. In particular,  $\theta_0$ help us understand the thermal structure and its relationship with physical, biological, and climatic processes. This approach is essential for modeling and predicting the evolution of marine systems in the Northeastern Atlantic \cite{ham2019deep}.

\begin{figure}[ht!] 
\centering 
\includegraphics[width=0.8\textwidth]{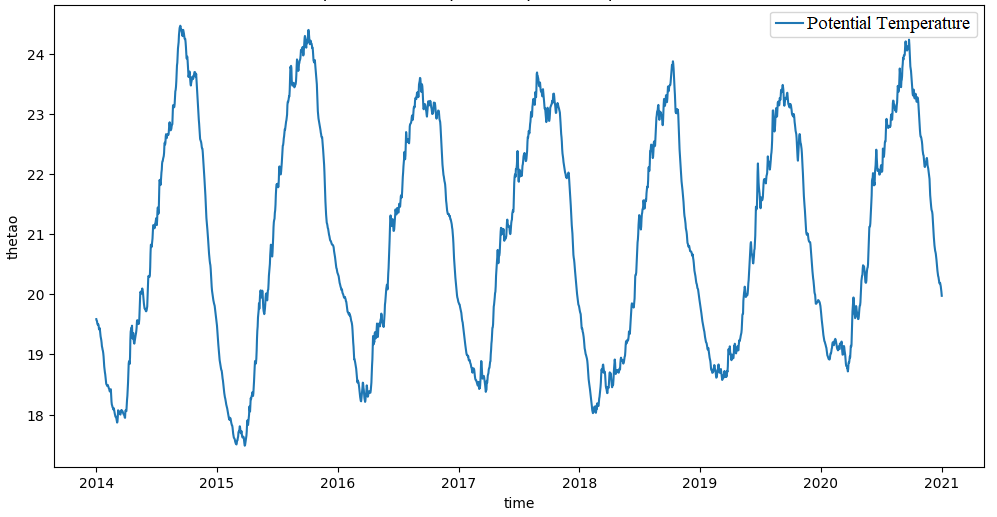} 
\caption{Temporal series of the study period for the $\theta_0$ variable.} 
\label{fig:serie_temporal_thetao} \end{figure}

Figure~\ref{fig:serie_temporal_thetao} shows a time series of potential temperature during the study period, elucidating seasonal variations.

\section{A Foundational Model for Forecasting the SST}
\label{se:method}

This section explains the Aurora model and how we adapted it to our purposes. It is a foundational model for atmospheric prediction, but its versatility permits it to be adapted to oceanography. It can process heterogeneous data with multiple variables, pressure levels, and spatial resolutions. These characteristics make Aurora a flexible and scalable system capable of forecasting the physical state of the ocean. This model can be adapted to more limited contexts with scarce data.

The scalability of this model is largely due to the local window attention mechanism and the U-Net~\cite{ronneberger2015unet} architecture. Both allow for information extraction at multiple scales. This multiscale capability is essential for capturing large- and small-scale structures. The model generates long-term forecasts by iteratively applying its prediction function in an autoregressive way:
\begin{equation}
\hat{X}^{t+1} = \Phi(X^{t},X^{t-1}),
\end{equation}
where $\hat{X}^{t+1}$ is the prediction at time $t+1$ and $\Phi$ is the model function. This scheme, analogous to recurrent networks, extends the temporal horizon, although it requires controlling the accumulation of errors. Aurora benefits from parallel training and mixed precision, reducing memory usage without degrading prediction quality. This facilitates its practical implementation in operational contexts, allowing for increased resolution without incurring prohibitive computational costs.

Aurora presents multiple configurations, differentiated by the number of parameters, the size of the embeddings, and the number of layers in the encoder and decoder. These variations impact computational performance and the model's ability to capture complex relationships.

\begin{table}[ht!]
\caption{Variants of the Aurora model.}
\centering
\begin{tabular}{|l|c|c|c|}
\hline
\textbf{Model} & \textbf{Layers Encoder/Decoder} & \textbf{Embedding Dimensions} & \textbf{Parameters in Millions} \\
\hline
Small Aurora & (2, 6, 2)/(2, 6, 2) & 256 & 117M \\
Medium Aurora & (6, 8, 8)/(6, 8, 8) & 384 & 660M \\
Large Aurora & (6, 10, 8)/(8, 10, 6) & 512 & 1.300M \\
\hline
\end{tabular}
\label{tab:modelos_aurora}
\end{table}

Table~\ref{tab:modelos_aurora} shows the size of the different configurations. We chose the medium-size model, looking for a balance between complexity and computational cost. Additionally, we adopted mixed precision to optimize memory.

The model's architecture consists of three components: the \textit{encoder}, the \textit{processor}, and the \textit{decoder}. The \textit{encoder} transforms raw inputs into a normalized internal representation, the \textit{processor} temporally evolves this representation, capturing significant relationships between the input data, and the \textit{decoder} restores the variables to their original resolution and levels. The architecture benefits from three-dimensional Transformers and attention mechanisms~\cite{vaswani2017attention, li2021learnablefourierfeaturesmultidimensional}, integrated with U-Net. These components allow feature extraction at various scales, processing data with different resolutions, levels, and static variables.

\subsection{The Encoder}
The objective of the encoder is to unify the inputs into a coherent three-dimensional representation. It converts heterogeneous data into a uniform internal representation, preparing it for the temporal and multiscale processing of the next module. It employs tokenization and compression of climatic states through cross-attention blocks based on a Perceiver IO~\cite{jaegle2021perceiverio} model. Thus, each atmospheric and surface variable is represented in a three-dimensional volume, incorporating Fourier encodings that add spatial and temporal information~\cite{li2021learnablefourierfeaturesmultidimensional}.

To normalize the information, embeddings based on sinusoidal encodings have been used, assigning each pressure level $p_l$ a representation of the type:
\begin{equation}
\text{Emb}(p_l) = \left[ \sin\left(\frac{p_l}{\lambda_i}\right), \cos\left(\frac{p_l}{\lambda_i}\right) \right].
\end{equation}

These embeddings integrate vertical information, reducing pressure levels to a fixed set of latent levels $C_L$ through Perceiver IO attention mechanisms. This allows the model to work with a manageable number of levels, maintaining the representativeness of vertical conditions while balancing computational cost. Dynamic and static variables are tokenized, incorporating positional and Fourier encoding, which unifies the differences between levels and resolutions.

\subsection{The Processor}
The backbone is the temporal core of the model, responsible for evolving the internal representation over time. To achieve this, it relies on a 3D Swin Transformer~\cite{liu2021swin, liu2022swin}, organized with a U-Net structure, which allows it to capture features at multiple scales.

An internal encoder gradually reduces spatial resolution and increases semantic depth. This is followed by an internal decoder, which recovers the original resolution and integrates the information obtained at each level. This hierarchical arrangement combines global and local information, capturing large-scale patterns and more detailed structures.

Attention is implemented through local windows, allowing each layer to process spatial subsets. This reduces computational complexity by avoiding global attention that spans entire layers. This local attention is given by:
\begin{equation}
\text{Attention}(\mathbf{Q},\mathbf{K},\mathbf{V}) = \text{Softmax}\left(\frac{\mathbf{Q}\mathbf{K}^\top}{\sqrt{d_k}}\right)\mathbf{V},
\end{equation}
where $\mathbf{Q}$, $\mathbf{K}$, and $\mathbf{V}$ are the query, key, and value matrices, respectively, and $d_k$ is the dimension of the keys. These local windows limit attention operations to specific regions. This strategy makes Aurora scalable, allowing it to handle higher resolutions without increasing computational cost.

The result is an enriched representation in which spatiotemporal relationships are explicitly modeled at multiple levels. This representation is then configured to be translated back to the original variables in the decoder.

\subsubsection{The Decoder}

The latent representation of the variables, generated by the processor, is restored to their original pressure levels and resolution in the decoder. This involves using inverse modules based on Perceiver IO models, which can reconstruct the desired variables from the internal representations, assigning each level and variable its corresponding spatial characteristics.

Each variable, $v$, and level, $l$, is obtained by applying a linear transformation and regrouping spatial patches until the original grid is recovered. Thus, the decoder ensures that the final predictions maintain coherence with the input dimensions and the information captured by the model is translated into physically interpretable forecasts.

\subsection{Subregional Oceanographic Prediction}

Adapting the model to oceanographic variables requires fundamental differences from atmospheric conditions. While Aurora was pre-trained assuming distributions and ranges typical of atmospheric variables (normalized in Kelvin and with pressure levels), oceanic variables, such as potential temperature, present narrower ranges and different scales. For example, sea surface temperature usually varies between approximately -2$^o$C and 30$^o$C, a narrower range than in atmospheric data. This disparity initially led to unrealistic predictions, as the model expected different distributions.

In the first step, we converted temperatures in Celsius to Kelvin and reduced depth levels. The data were normalized to zero mean and unit variance. Geographic coordinates were refined to maintain consistency in the latitude-longitude grid, and the land mask was carefully interpolated to ensure seamless integration of oceanic variables within the model's infrastructure.

\begin{figure}[ht!]
    \centering
    \includegraphics[width=0.9\textwidth]{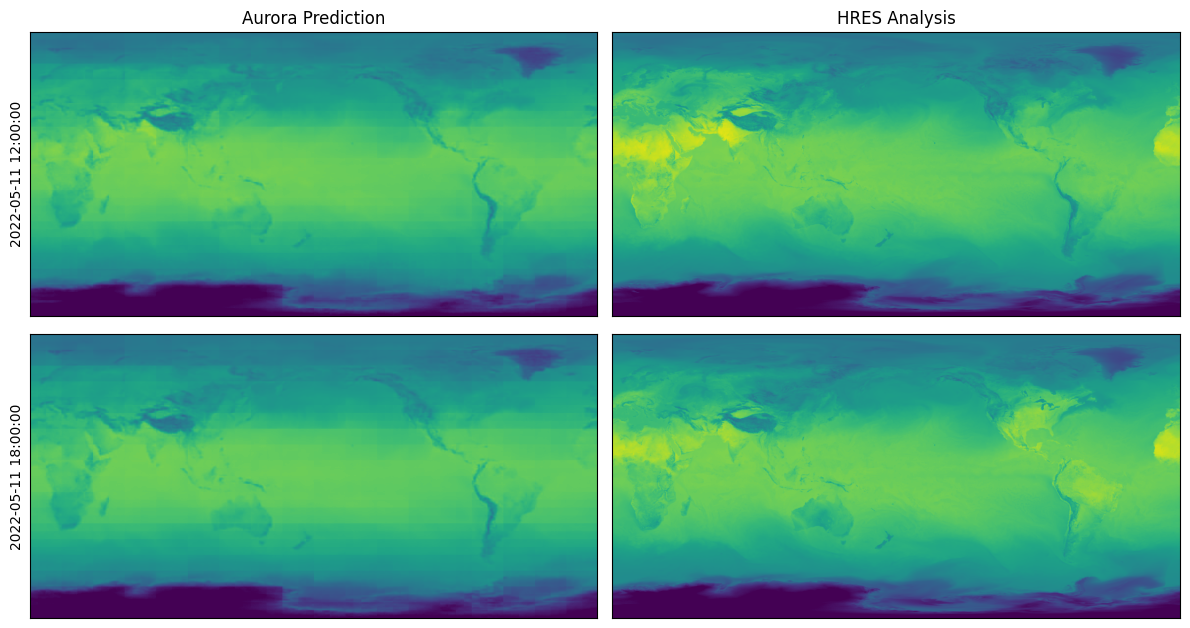}
    \caption{Prediction of the pre-trained Aurora model with meteorological data after reducing data resolution to $0.5^\circ$.}
    \label{fig:test_0}
\end{figure}

We scaled the original atmospheric data to a lower resolution of $0.5\circ$ degrees and carried out an inference with the pre-trained Aurora model. Figure~\ref{fig:test_0} shows the global temperature map in Kelvin in two different time instants and compares to the HRES product. These results indicate that the resolution change was successful. 

We adapted the oceanographic data by interpolating, filling in missing values, and converting to Kelvin. The latitude and longitude coordinates are interpolated to match the model's resolution, and missing values are replaced with the mean. The pre-trained model is loaded, and the predictions with our data are consistent with the expected global atmospheric patterns.

Initial attempts with oceanic data did not yield good performance, as the model still carried a bias inherited from its pre-training. To overcome this challenge, we fine-tuned the model in two phases. In the first phase, the entire network was frozen except for the decoder, so that it could learn the ocean variables without altering the already acquired internal representations. Once the decoder assimilated the new dynamics, all the model's parameters were trained using a lower learning rate to avoid losing prior knowledge. We used the AdamW optimizer as in the original work~\cite{bodnar2024aurora}, which helps regularize the model and avoids overfitting.

\section{Experimental Configuration}
\label{se:experimental_settup}

\subsection{Dataset Split}

The split of the dataset for training the neural network considered oceanic conditions, such as seasonal cycles or adverse phenomena. For this reason, the training set was defined annually between 2014 and 2018, the validation set comprised 2019, and data from 2020 was reserved for the test set. This roughly represents 70\% for training, 15\% for validation, and 15\% for testing, maintaining the temporal order to prevent overlaps. This approach ensures that data used to validate and test the model are independent of those used in training, for assessing its ability to make predictions in unobserved scenarios. 

Splitting data in this way also benefits the study of oceanic phenomena, as training with data from an extended period allows the model to process multiple conditions. Additionally, using recent data for validation and testing, evaluates the model's adaptability to current conditions. This classification also ensures that each subset contains samples from all seasons.  

\begin{figure}[ht!]
\centering
\includegraphics[width=0.7\textwidth]{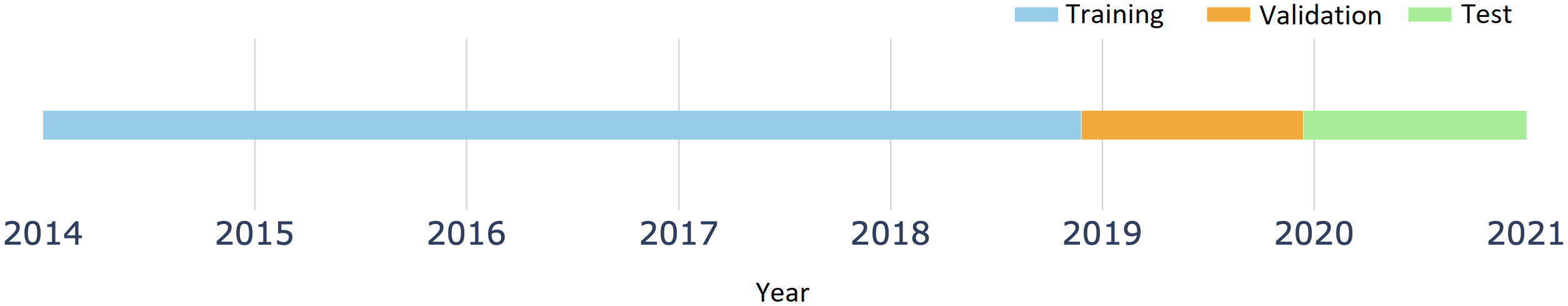}
\caption{Dataset split into the training, validation, and test sets.}
\label{fig:division_temporal}
\end{figure}

Figure~\ref{fig:division_temporal} illustrates how the data was divided. To attain the 70\%-15\%-15\% rate, the training set spans from January 1, 2014, to November 25, 2018; the validation set covers November 26, 2018, to December 13, 2019; and the test set extends from December 14, 2019, to January 1, 2021.

Batches are organized with samples of three consecutive values in sliding windows over the temporal axis. The samples are shuffled, and a padding mechanism is applied to ensure that all batches contain the same number of samples. The process begins with the generation of sliding windows, a mechanism that divides the dataset into consecutive temporal partitions. The window size is adjusted depending on the number of days in each window.  

\subsection{Loss Function}

We used the mean absolute error (MAE) as the loss function for optimizing the model's parameters. Its formula is given by:
\begin{equation}
\text{MAE} = \frac{1}{N}\sum_{i=1}^N |y_i - \hat{y}_i|,   
\end{equation}
where $y_i$ is the ground truth value, $\hat{y}_i$ is the model prediction, and $N$ the total number of predictions. In case of multiple variables, a loss function similar to MAE could be employed. The function below is a variant of MAE that takes into account the sum across variables with their respective weights and scales:

\begin{align}
\mathcal{L}(\hat{X}^t, X^t) = & \frac{\gamma}{V_S + V_A}
\left[
\alpha \left(
\sum_{k=1}^{V_S} \frac{w_k^S}{H \times W} \sum_{i=1}^{H}\sum_{j=1}^{W} |\hat{S}_{k,i,j}^t - S_{k,i,j}^t|
\right)
\right.
 \nonumber \\
& \left. + \beta \left(
\sum_{k=1}^{V_A}\frac{1}{C \times H \times W}\sum_{c=1}^C w_{k,c}^A \sum_{i=1}^{H}\sum_{j=1}^{W} |\hat{A}_{k,c,i,j}^t - A_{k,c,i,j}^t|
\right)
\right],
\end{align}
where \(\hat{X}^t = (\hat{S}^t, \hat{A}^t)\) is the predicted state, and \(X^t = (S^t, A^t)\) is the actual state, with \(S\) and \(A\) representing surface and atmospheric variables, respectively. Here, \(\alpha\), \(\beta\), and \(\gamma\) are weighting factors; \(w_k^S\) and \(w_{k,c}^A\) are the weights for each variable; and \(H, W, C, V_S, V_A\) are the spatial dimensions and the variable set. The loss function adjusts the contribution from each variable, prioritizing the most relevant for the study's objective.

\subsection{Evaluation Metrics}

The metrics used to evaluate the model's performance were the root mean squared error (RMSE), the weighted bias (BIAS), and the anomaly correlation coefficient (ACC)~\cite{ben2024rise}. These metrics consider weights based on the cosine of latitude to account for the Earth's spherical geometry.   

\begin{figure}[ht!]
 \centering
    \includegraphics[width=0.7\textwidth]{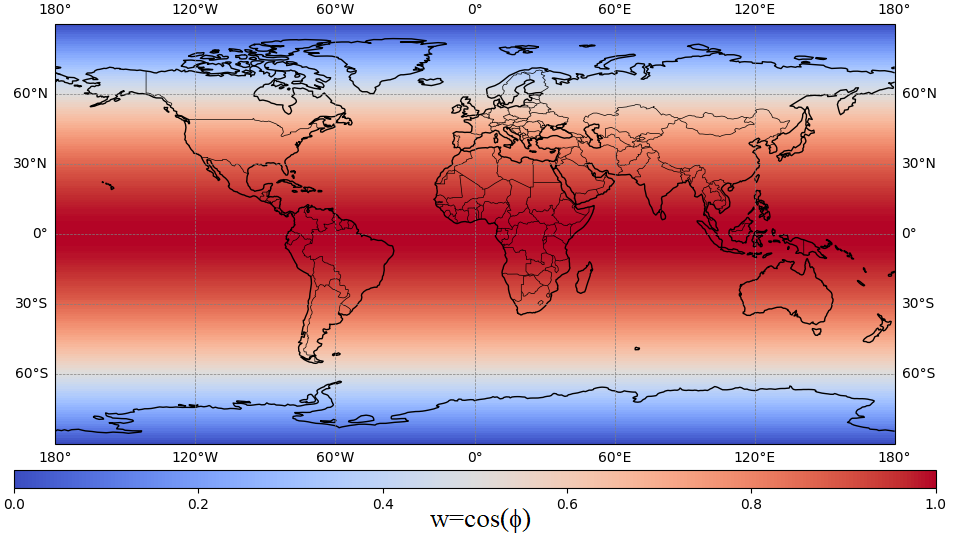}
    \caption{Latitud-based weights used in the metrics to account for the size of the patches induced by the Earth's spherical geometry.}
    \label{fig:pesos_latitud}
\end{figure}

Figure \ref{fig:pesos_latitud} illustrates the weights, defined as $w = \cos(\phi)$, which decrease toward the poles. This ensures that the weighted metrics remain globally representative.

The weighted RMSE calculates the average magnitude of the error between the predictions $\hat{y}_i$ and the ground truth values $y_i$ and is computed as:  
\[
\text{RMSE} = \sqrt{\frac{1}{N}\sum_{i=1}^N w_i^\text{norm} (y_i - \hat{y}_i)^2},
\]
where $w_i^\text{norm}$ is defined as:
\[
w_i^\text{norm} = \frac{w_i}{\overline{w}} = \frac{\cos(\phi_i)}{\overline{w}}, \; \text{with} \; \overline{w} = \frac{1}{N} \sum_{j=1}^N w_j.
\]

In this formulation, $w_i$ represents the original weight associated with each observation, and $\overline{w}$ is the average of these weights. This approach ensures that the error is not biased toward regions with higher data density.  

The weighted Bias was used to evaluate the average difference between predictions and actual values. Its calculation helped identify whether the model systematically underestimates or overestimates observed values. The formula is given by:  
\[
\text{Bias} = \frac{1}{N}\sum_{i=1}^N w_i^\text{norm} (y_i - \hat{y}_i).
\]

This measure provides an accurate metric for identifying the model’s systematic tendencies. The weighted ACC, on the other hand, measures the model’s ability to replicate variation patterns between predictions and actual values. It is calculated as
\[
\text{ACC} = \frac{\text{Cov}(A, B)}{\sqrt{\text{Var}(A) \cdot \text{Var}(B)}},
\]
where $\text{Cov}(A, B)$ is the covariance and $\text{Var}(\cdot)$ the variance, given by
\begin{align*}
    \text{Cov}(A, B) & = \frac{1}{N}\sum_{i=1}^N w_i^\text{norm} A_i B_i, \\
    \text{Var}(A) & = \frac{1}{N}\sum_{i=1}^N w_i^\text{norm} A_i^2, \\
    \text{Var}(B) & = \frac{1}{N}\sum_{i=1}^N w_i^\text{norm} B_i^2.
\end{align*}
with $A_i = y_i - \overline{y}$ and $B_i = \hat{y}_i - \overline{\hat{y}}$ the anomalies, i.e., the deviations from the weighted averages, and
\[
\overline{y} = \frac{1}{N}\sum_{i=1}^N w_i^\text{norm} y_i, \quad
\overline{\hat{y}} = \frac{1}{N}\sum_{i=1}^N w_i^\text{norm} \hat{y}_i.
\]

This process ensures that the ACC reflects the model's ability to capture global patterns over the prediction anomalies compared to the actual values.

\section{Results}
\label{se:results}

We used the test set to evaluate the model. In the first experiment, we assess the performance of the fine-tuning strategy. We follow three strategies: in the first one, we unfroze all the parameters of the network and trained with a low learning rate of 1$\times10^{-5}$; in the second one, we only unfroze the decoder and trained with a learning rate of 1$\times10^{-4}$, so that atmospheric knowledge was maintained in the backbone, and oceanic features were gradually learned by the decoder; in the last one, we maintained de parameters of the previous strategy and unfroze all the layers, training the model with a low learning rate of 1$\times10^{-5}$ and a few epochs.

We used a batch size of 3 and 15 epochs, obtaining an RMSE of 0.131K for the first strategy and 0.140K for the second strategy. Finally, the network of the latter was fully fine-tuned with a learning rate of 1$\times10^{-5}$, obtaining an RMSE of 0.124K.

These experiments were repeated, increasing the batch size to 8 samples and using more computational resources. This contributed to an approximate time reduction of 15\% per epoch. We also increased the number of epochs to 30 and obtained an RMSE of 0.130K for the first strategy and 0.135K for the second one. The last fine-tuning step reduced the RMSE to 0.134K. Table ~\ref{tab:resultados_metricas} shows the final results obtained in these experiments. 

\begin{table}[ht!]
\caption{Results of the fine-tuning process. In the first row, the network was completely fine-tuned using a small learning rate. In the second row, only the decoder was fine-tuned with a larger learning rate. In the third row, the resulting network was subsequently fine-tuned with a reduced learning rate. In the last three rows, we show the same experiments increasing the batch size to 8 and the number of epochs to 30. Bold letters represent the best results in each column.}
\centering
\begin{tabular}{lccc}
\toprule
\textbf{Setting} & \textbf{RMSE (K)} & \textbf{Bias (K)} & \textbf{ACC} \\
\midrule
Full fine-tuning, $lr=1\times10^{-5}$, batch=3     & 0.131 & -0.069 & 0.997 \\
Decoder fine-tuning, $lr=1\times10^{-4}$, batch=3 & 0.140 & -0.064 & 0.997 \\
Full fine-tuning with new decoder, $lr=1\times10^{-5}$, batch=3 & \textbf{0.124} & -0.064 & 0.997 \\
Full fine-tuning, $lr=1\times10^{-5}$, batch=8     & 0.130 & -0.062 & 0.997 \\
Decoder fine-tuning, $lr=1\times10^{-4}$, batch=8 & 0.135 & -0.059 & 0.997 \\
Full fine-tuning with new decoder, $lr=1\times10^{-5}$, batch=8 & 0.134 & \textbf{-0.033} & 0.997 \\
\bottomrule
\end{tabular}
\label{tab:resultados_metricas}
\end{table}

When the batch size was increased to 8, the results became somewhat more competitive. This finding shows that with better resources, using larger batches speeds up training and also improves the quality of the predictions. The high ACC (ACC $\approx 0.997$) suggests that the model captures the relationships between actual and predicted anomalies. This agreement reveals that the model does not simply fit the mean of the data but also accounts for unique phenomena. As for the Bias, in most cases, values close to -0.060K were obtained, except for the last training, where we obtained a lower bias of -0.033K. 

\begin{figure}[ht!]
\centering
\includegraphics[width=\textwidth]{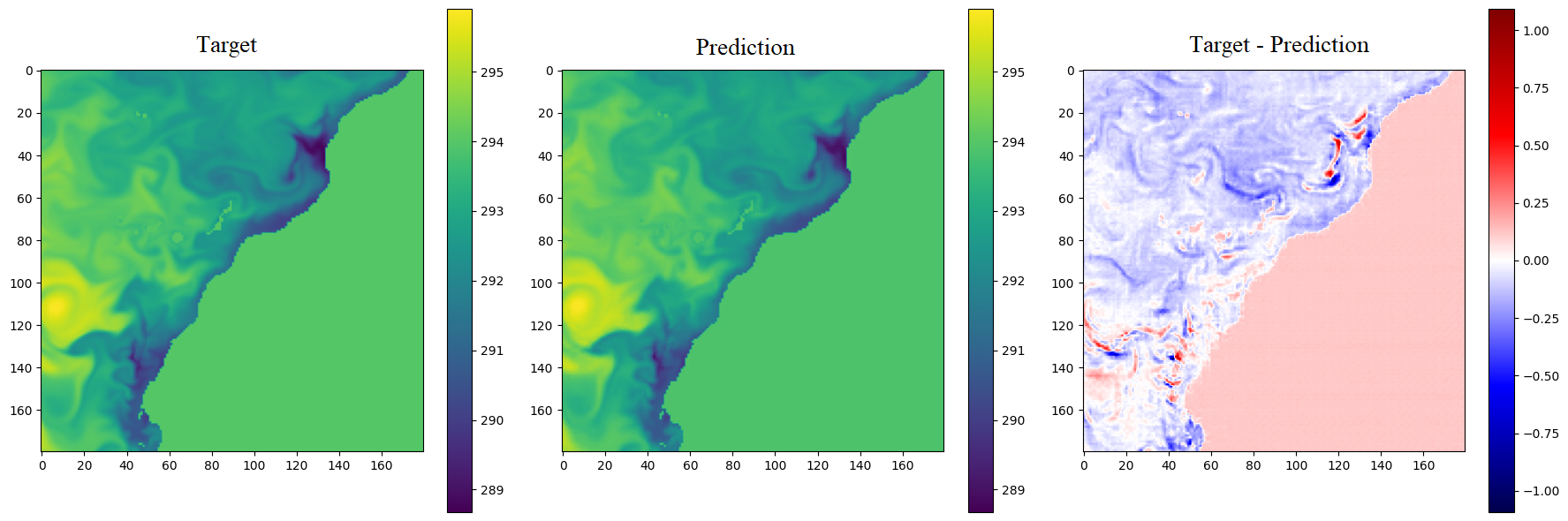}
\caption{Error produced in a forecast of the potential temperature in Kelvin for one-day prediction.}
\label{fig:predicciones}
\end{figure}

Figure~\ref{fig:predicciones} presents an example of an inference. This figure shows the target value, the model prediction, and the difference between the two. The values are represented in Kelvin with color scales, where greenish tones indicate intermediate values, and blue and yellow highlight more extreme values between the prediction and target values. The difference between them is shown with a scale centered on zero, where red and blue indicate overestimation and underestimation, respectively.

It reveals that the potential temperature prediction model has more errors in certain regions, such as in coastal zones. This is due to the inherent complexity of these areas, characterized by more intense temperature variations compared to other zones, as noted in~\cite{vazquez2021assessment}.

The difference manifests the variability in coastal zones through more saturated tones, indicating that the model struggles at the ocean-continent boundary. This difficulty is likely due to the interaction of various factors influencing potential temperature in these zones, such as ocean currents, coastal topography, and local winds. These factors can generate microclimates and significant temperature variations at small spatial scales.

In contrast, conditions in the open ocean are more uniform and stable, and the difference between the model prediction and the observed data is reduced. The lower variability in these areas allows for capturing potential temperature patterns with greater accuracy.
This suggests that specific training aimed at improving the results in coastal areas could improve the overall model's performance. This training could incorporate high-resolution data of relevant variables in coastal zones or use modeling techniques that better capture the complexity of these zones.

\begin{figure}[ht!]
\centering
\includegraphics[width=\textwidth]{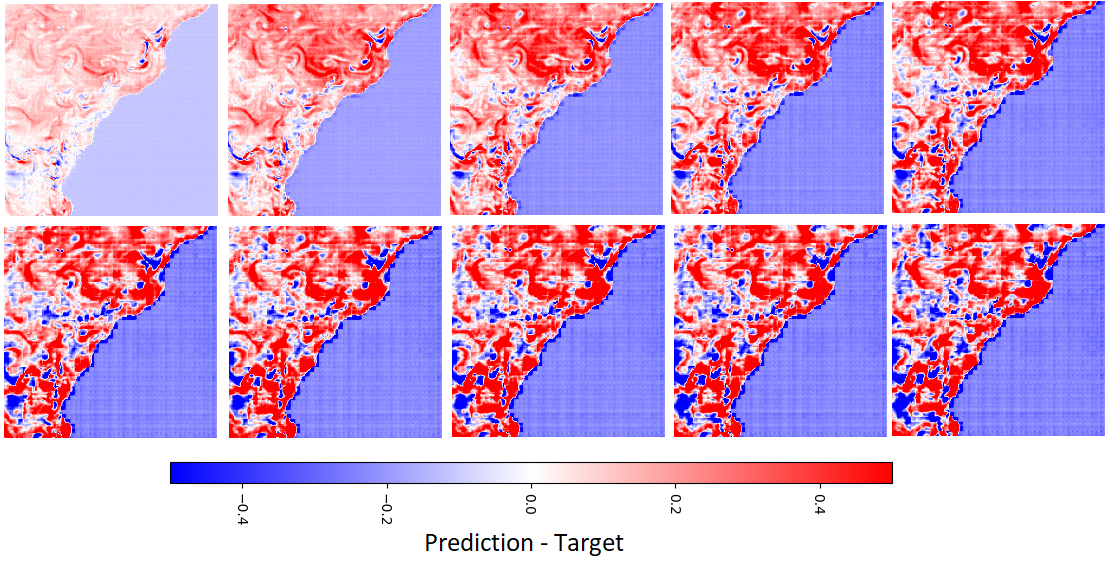}
\caption{RMSE error in autoregressive ten-day forecasts.}
\label{fig:rollout_evolucion}
\end{figure}

An additional experiment consisted of increasing the prediction to ten days. Each sample included ten temporal steps, and the data was analyzed at each lead time. As the time horizon increases, we arrive at an increasingly challenging scenario, as uncertainty accumulates and small initial inaccuracies can be amplified. Figure~\ref{fig:rollout_evolucion} shows the evolution of the difference between the prediction and the target value for ten-day forecasts. The difference is small in the first days, and errors accumulate with more autoregressive steps, which become accentuated in the last days. This figure is particularly interesting for identifying errors and analyzing problematic regions. This long-term visualization demonstrates that maintaining effectiveness with an increased time horizon presents a significant challenge. 

\begin{figure}[ht!]
\centering
\includegraphics[width=0.6\textwidth]{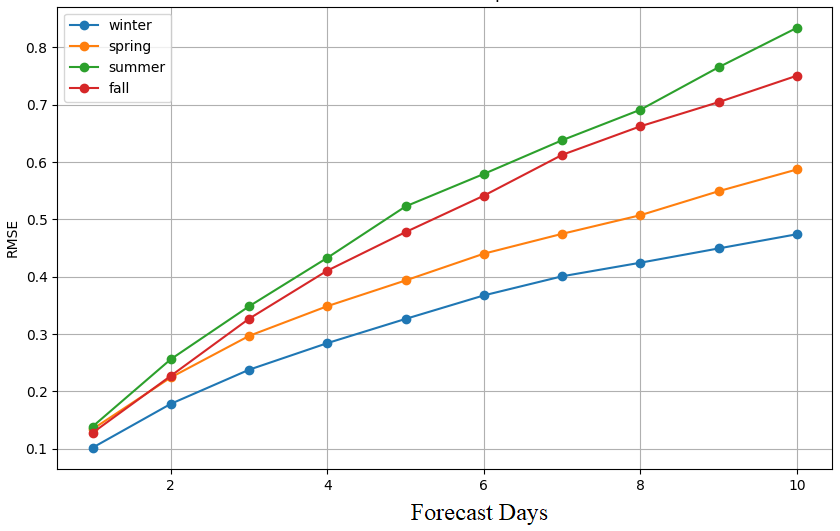}
\caption{Seasonal average RMSE for ten lead times.}
\label{fig:rollout_estaciones}
\end{figure}

Figure~\ref{fig:rollout_estaciones} shows the average RMSE for the four seasons over ten-day forecasts. The RMSE increases progressively as the time horizon expands, with greater increases in those seasons that present more difficult conditions. Indeed, summer (plotted in green) shows a more pronounced increase, reaching values close to 0.9K after 10 days, suggesting that seasonal cycles significantly influence the results.

\section{Discussion}
\label{se:discussion}

Oceanography has traditionally relied on physically based numerical models. Although these models are robust and grounded in well-established physical principles, they are often complex, computationally expensive, and limited in representing certain spatial and temporal scales. In contrast, deep learning models derive relationships directly from data, operating without explicit physical constraints. They can anticipate phenomena that are difficult to reproduce numerically and, with sufficient data availability, have the potential to accelerate forecasting processes~\cite{etoffs2022challenges}.

The results presented in this study demonstrate that deep learning models can leverage prior knowledge from foundational models trained in different contexts. Oceanographic research can benefit from these advances through reduced computational costs, easier integration of diverse data sources, rapid generation of forecasts, and the ability to test numerous configurations efficiently. Given the strong dependence of oceanography on data quality and quantity, the capability of deep learning techniques to assimilate and exploit vast datasets paves the way for long-term innovation.

This work expands the range of forecasting methodologies available to oceanography, introducing approaches not previously explored in the field~\cite{etoffs2022challenges}. The use of foundational models pretrained in other fields helps lower barriers to entry and enhances resource efficiency.

Despite these advantages, several limitations remain: Training three-dimensional models is computationally demanding; hardware constraints significantly affected this study, as each training session required considerable time, limiting both resolution and the number of variables; addressing this challenge would require more powerful resources as suggested in~\cite{aurora_finetuning}.

Another key challenge concerns data representativeness. Although the dataset employed here is based on reanalysis products, these are derived from numerical models and partial observations. Consequently, predictions may inherit biases from the underlying data. Regions with sparse observations or suboptimal conditions may yield less accurate results. Future work should incorporate additional information, such as \textit{in situ} measurements and satellite data, to better evaluate and improve prediction quality.

This study focuses solely on potential temperature; however, ocean dynamics depend on multiple interacting variables, including salinity, density, and currents. Extending the model to incorporate these factors will increase complexity, as the model must learn new interdependencies. Nevertheless, this added complexity could yield more realistic oceanic scenarios and improve predictive accuracy.

Another limitation lies in model interpretability. Neural networks remain largely opaque. Future research should therefore investigate the internal representations and decision-making processes of these models. Enhancing explainability will foster a deeper understanding of how predictions are generated and allow for greater confidence and control~\cite{reichstein2019deep}.

Despite these challenges, this work opens promising avenues for future exploration. Integrating larger and higher-quality datasets could extend prediction horizons and enhance model robustness. Including additional variables, such as salinity and current velocity, would enable the study of hydrodynamic processes and provide a more comprehensive understanding of ocean behavior. Increasing spatial resolution through high-resolution models is another promising direction, though it will demand greater computational power. This could be achieved through progressive training strategies or by dividing the domain into smaller subregions for subsequent integration.

Finally, incorporating physical knowledge directly into training represents another exciting research direction. Physics-informed neural networks (PINNs) introduce physical constraints by penalizing violations of conservation laws in the loss function. Embedding simplified fluid dynamics equations within the learning process can reduce the model’s dependence on data alone, guiding it toward physically consistent solutions~\cite{raissi2019physics, cai2021pinns}.

Continual learning also presents a valuable opportunity. Because the ocean is a dynamic and ever-changing system, models that can be regularly updated with new data will improve adaptability and responsiveness to evolving conditions. Such models could react more effectively to extreme events and provide continuously improving forecasts over time.

\section{Conclusion}
\label{se:conclusion}

This study established the conceptual, methodological, and technical groundwork for adapting a foundational atmospheric model to oceanographic applications. Specifically, the Aurora model was successfully applied to forecast ocean potential temperature in a local region.

Although the atmosphere and ocean are tightly coupled components of the climate system, they differ in structure and dynamics. Atmospheric variables---on which Aurora was originally trained---are organized by pressure levels and evolve rapidly, whereas oceanic variables are distributed by depth, governed by thermo-haline properties, and change more slowly due to higher density~\cite{talley2011descriptive}. Aurora’s capacity to assimilate oceanic data and produce coherent results demonstrates cross-domain adaptability, suggesting potential for future knowledge transfer between disciplines.

While this study focused on potential temperature, the experiments indicate that Aurora can be extended to more complex, multivariable scenarios given sufficient computational resources. The sensitivity observed across training configurations, such as partial layer freezing and learning rate adjustments, highlights the importance of hyperparameter tuning and normalization strategies in achieving optimal performance. Overall, Aurora’s pre-trained parameters can be effectively fine-tuned for new oceanographic tasks, reinforcing the model’s flexibility for broader environmental applications.

\bibliographystyle{unsrt}  
\bibliography{references}  

\end{document}